# Airport Taxi Time Prediction and Alerting: A Convolutional Neural Network Approach

Erik Vargo[1], Alex Tien[2], and Arian Jafari[3]
*The MITRE Corporation, McLean, VA, 20102, U.S.A.*

**This paper proposes a novel approach to predict and determine whether the average taxi-out time at an airport will exceed a pre-defined threshold within the next hour of operations. Prior work in this domain has focused exclusively on predicting taxi-out times on a flight-by-flight basis, which requires significant efforts and data on modeling taxiing activities from gates to runways. Learning directly from surface radar information with minimal processing, a computer vision-based model is proposed that incorporates airport surface data in such a way that adaptation-specific information (e.g., runway configuration, the state of aircraft in the taxiing process) is inferred implicitly and automatically by Artificial Intelligence (AI).**

## I. Introduction

Taxi-out time is an indicator of departure efficiency and is often the early signal of large holding and diversion events for airports that have constrained surface space. This is one of the real-time performance metrics that is of great interest to air traffic managers and flight dispatchers. For a busy airport that has limited tarmac space like LaGuardia Airport (LGA), an increasing average taxi-out time under a deteriorating visibility condition could soon lead to surface gridlock that would cause significant delays to both arrivals and departures. Thus, research is needed to develop an early alert or prediction of long taxi-out times to enable early delay mitigation actions.

The problem of predicting taxi-out times has received considerable treatment in the aviation literature. Most research exploring the domain of taxi time prediction has focused on predicting taxi-out times for individual aircraft. Inaccurate taxi-out times can lead to a variety of National Airspace System (NAS) inefficiencies, such as a reduction in predictability for downstream Traffic Flow Management (TFM) applications and excess fuel consumption after push back from the gate. By better predicting aircraft-specific taxi-out times, informed updates can be made to the flight schedule to improve predictability and more efficiently use available NAS resources (e.g., capacity). Although our focus is on predicting average taxi-out time, it's worth reviewing the literature on aircraft-specific taxi-out time predictions for historical context.

Prior to the availability of Airport Surface Detection Equipment, Model X (ASDE-X) data, researchers leveraged non-positional aircraft data to calibrate taxi-out time prediction models. Combining subject matter expertise and statistical analysis, features were selected that demonstrated high correlation with taxi-out times, such as departure queue size, weather impacts (e.g., Severe Weather Avoidance Plan [SWAP]) and related downstream restrictions, departure gate, and runway configuration. While early efforts were limited to classical linear regression [1], [2], more recent work has applied modern machine learning (ML) algorithms, such as support vector machines, random forests, and neural networks, albeit to analogous feature data [3], [4], [5].

With the availability of ASDE-X surface data via the Federal Aviation Administration's (FAA) System Wide Information Management (SWIM) feed, researchers can now construct a more granular and dynamic view of tarmac operations. Because ASDE-X provides position updates that are both near real-time and attributable to distinct aircraft, an aircraft's path from departure gate to take off can be tracked with high accuracy and partitioned among the various stages of its taxiing procedure, e.g., push back, taxiing to the runway queue, takeoff. A taxi-out time prediction model can likewise be partitioned by stage, resulting in a more operationally nuanced and accurate predictions. Noting the high correlation between departure queue length and taxi-out time, [6] use ASDE-X data to estimate a "virtual departure queue," which is subsequently used to predict taxi-out time via a linear regression. [7] uses ASDE-X data to split the taxi-out time prediction into two components: the first is defined from push back to arrival at the runway queue, and the second is defined from the runway queue to takeoff. This splitting is only possible due to the real-time and granular nature of the ASDE-X data feed.

---

[1] Principal Artificial Intelligence Engineer, Operational Performance Department, AIAA Member.
[2] Principal Artificial Intelligence Engineer, Operational Performance Department, AIAA Member.
[3] Senior Artificial Intelligence Engineer, Operational Performance Department.

Broadly speaking, existing taxi-out prediction models can be categorized as: 1) data-driven ML algorithms that use higher-level, static (i.e., not updated in real-time) aircraft metadata to establish correlations with taxi-out times, and 2) adaptation-driven algorithms that lean heavily on the dynamic view offered by ASDE-X, coupled with more classical regression models, to improve accuracy. It is difficult to say which category represents the "state of the art," as studies tend to focus on different sets of airports and apply different metrics for evaluation, often over different time ranges. Despite this challenge, the strengths and weaknesses of each approach are clear. ML approaches are more extensible, flexible, and transferrable because models can be trained somewhat analogously across multiple airports with little change to the underlying architecture, albeit at the expense of operational sensitivity. Adaptation-driven models have the advantage of greater operational nuance, although the time required to build adaptations may offset the gains in accuracy, especially if adaptations must be routinely updated to keep up with operational changes at the airport.

In this paper, we propose an approach that attempts to combine the strengths of both approaches: a computer vision-based ML model that incorporates ASDE-X surface data in such a way that adaptation-specific information (e.g., runway configuration, the state of aircraft in the taxiing process) is inferred implicitly and automatically by Artificial Intelligence (AI). As a result, we reap both the extensibility benefits of ML operational benefits of ASDE-X. In the following sections, we more formally define both the taxi-out prediction problem and the computer vision model developed to solve it.

## II. Problem Statement

Our goal is to develop a model to accurately determine whether the average taxi-out time at an airport will exceed a pre-defined threshold within the next hour of operations. Prior work in this domain has focused exclusively on predicting taxi-out times on a flight-by-flight basis. Our focus on the average taxi-out time metric has several advantages. First, while aircraft-specific taxi-out times are more relevant for downstream scheduling, individual taxi-out times do not explicitly indicate whether current operations as a whole are trending in a favorable or unfavorable direction. As such, they do not provide the operator with a clear indication of whether tactical action is necessary to mitigate congestion on the tarmac. Second, deriving aggregate predictions from the outputs of a model tuned to predict individual taxi-out times will exhibit greater error on the derived (aggregated) prediction than a model constructed to predict the aggregate metric as its primary goal. Finally, aircraft-specific taxi-out time predictions are limited to aircraft that are currently on the tarmac. In contrast, the average taxi-out time is a function of both current taxiing aircraft and arrivals that have yet to land and may soon contribute to the aggregate. The computer vision model we propose below fuses ASDE-X data from taxiing aircraft with projected arrival throughput statistics to better handle this interaction.

## III. Data Sources

In this section we define the data sources that make up our model's feature set, which include tarmac "images" and additional tarmac metadata.

### A. Tarmac Images

Arrival and departure images are generated from ASDE-X data at a rate of one image per minute, although higher frequency updates are possible and will be explored in the future. Each image is constructed by first gridding the target airport's tarmac region at a resolution that's granular enough so that aircraft occupying distinct locations on the tarmac can be mapped to distinct bins within the grid. The extent of the airport's grid is informed by an exploration of historical ASDE-X hits on the tarmac; these hits are visualized by a heatmap, and an extent is chosen to bound the tarmac such that most historical hits fall within the bounds.

As an example, Figure 1 illustrates this process for LGA, which includes an initial exploratory heatmap (left) and the chosen (right). The LGA grid consists of 20 bins and 33 bins in the latitude and longitude dimensions, respectively. In general, the number of bins in the latitude dimension (H) and longitude dimension (W) are a function of the range of the extent in each direction.

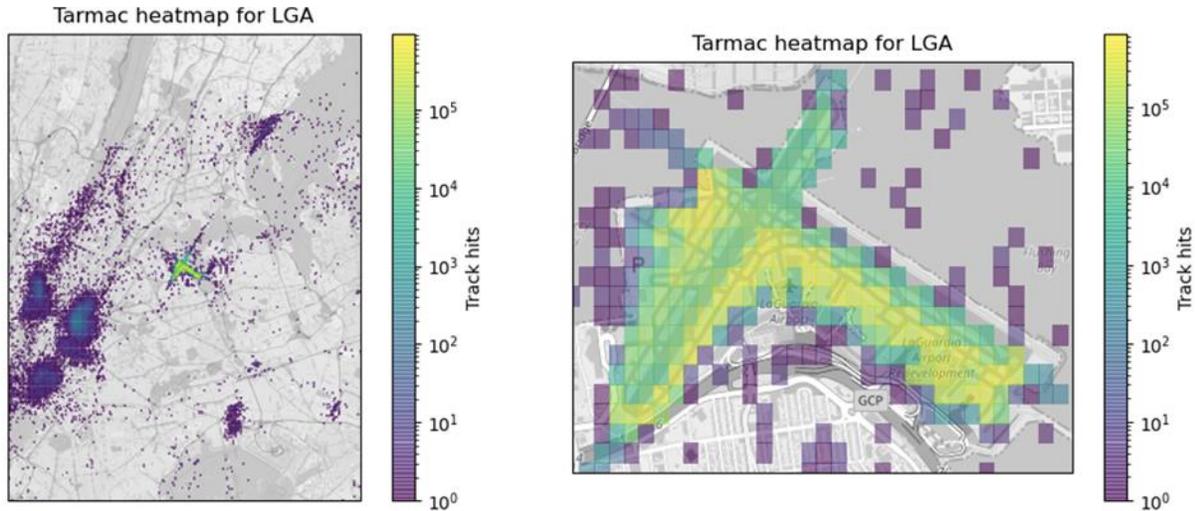

**Figure 1. Initial (left) and Cropped (right) Heatmaps of LGA ASDE-X Hits**

Once the grid is defined for an airport for any one-minute slice of time, we parse the ASDE-X data—pulled from the FAA's System-Wide Information Management (SWIM) feed—and extract each arrival flight's current location on the grid. We repeat this process for departure flights. A digitized view of tarmac arrivals is then constructed by populating an H×W array with 1s at indices where arrivals are present, and 0s elsewhere. Departures are coded analogously in a separate channel of the image array; by logically separating arrivals and departures, the algorithm can more easily attribute impact as a function of each aircraft's taxiing phase. See Figure 2 for an illustration, where a surface snapshot of taxiing flights at LGA on 1 September 2015 at 0Z (left) is converted into an array representation (right) where bins with arrivals are colored blue and bins with departures are colored red.

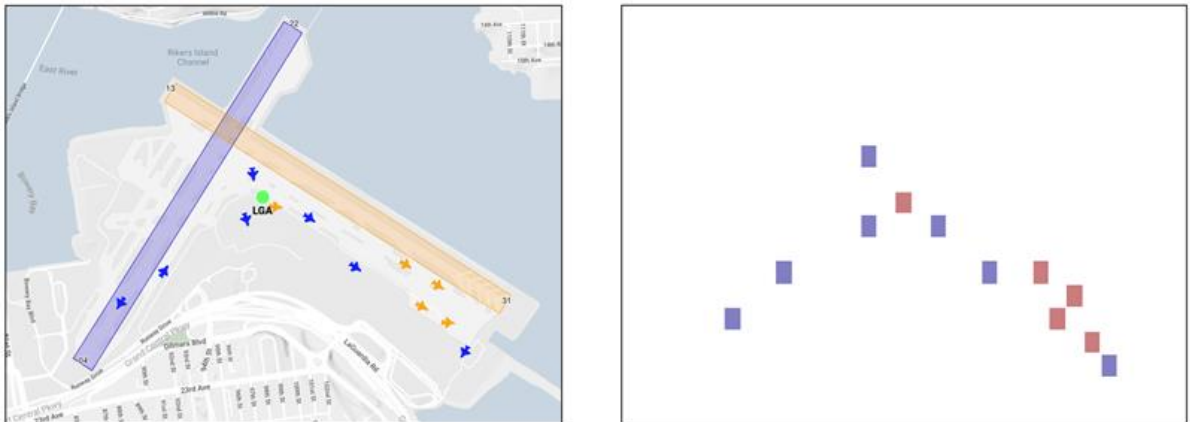

**Figure 2. NOD Snapshot of LGA on 1 September 2015 at 0Z (Left) and Array Representation (Right)**

The above process creates arrival and departure images that contain a single, binary valued channel for each entry in their respective arrays. Also, at each array entry (i,j) where the binary channel contains a 1 (i.e., an aircraft is present), we include two additional channels containing the aircraft's current ground speed and cumulative taxi time (taxi-in or taxi-out, depending on whether the grid cell is occupied by an arrival or a departure). In total, there are six channels of information at each one-minute time slice. Note that any given sample from our dataset consists of 30 such arrays in sequence. The first array is computed 29 minutes in the past and the final array is computed at the present time.

### B. Tarmac Metadata

We supplement these aircraft-specific features with aggregate features to better capture higher level trends in tarmac operations and performance. These features include the previous 30 minutes of the following statistics, which can be derived from the SWIM feed:

- Average taxi-in time: The average taxi-in time computed with respect to all aircraft currently on the tarmac.
- Average taxi-out time: The average taxi-out time computed with respect to all aircraft currently on the tarmac.
- Current arrival demand: Arrival demand for the current hour, frozen at the top of the hour
- Current departure demand: Departure demand for the current hour, frozen at the top of the hour
- Current arrivals: Number of arrivals since the top of the hour, updated each minute
- Current departures: Number of departures since the top of the hour, updated each minute
- Arrival demand (+1 hour, +2 hour): The projected arrival demand over the next one and two hours, starting from the top of those hours. These projections can change every minute as new flight data becomes available.
- Departure demand (+1 hour, +2 hour): The projected departure demand over the next one and two hours, starting from the top of those hours. These projections can change every minute as new flight data becomes available.

The tarmac metadata features are, collectively, a time-series of length 30. Slices are one-minute separated—starting from 29 minutes in the past—and each slice contains an eight-dimensional (8D) vector of the above statistics.

## IV. Model Construction

The ML model consists of two sets of input features, which are first processed by independent neural networks, before being fused together prior to generating the final binary prediction of whether average taxi-out time will exceed the threshold within the next one-hour. Figure 3 illustrates the deep neural network architecture. The length-30 sequence of tarmac images, which includes channels defining arrival and departure flight locations, along with ground speeds and cumulative taxi-in/taxi-out times, is fed into a three-dimensional (3D) Convolutional Neural Network (CNN) or Conv3D. Each dimension of the convolution operates on a separate dimension of the "video." The first two dimensions of the convolution slide a 2D window across all channels of each time-sliced image, extracting features and compressing the input in the process. The final dimension of the 3D convolutions extends the window across the time domain with the goal of further compressing the input sequence and adding a temporal component to feature extraction. Note that while a recurrent layers such as a convolutional LSTMs (Long Short-term Memory) could likewise be used to analyze a time-series of images, we found that the computational burden of recurrent architectures – compared to the Conv3D model – more than offset gains in accuracy.

The length-30 sequence of tarmac metadata is likewise fed through a one-dimensional (1D) CNN (or Conv1D). In this case, the sliding window operates on the input's temporal dimension, and the 8 feature dimensions act as 8 separate data channels.

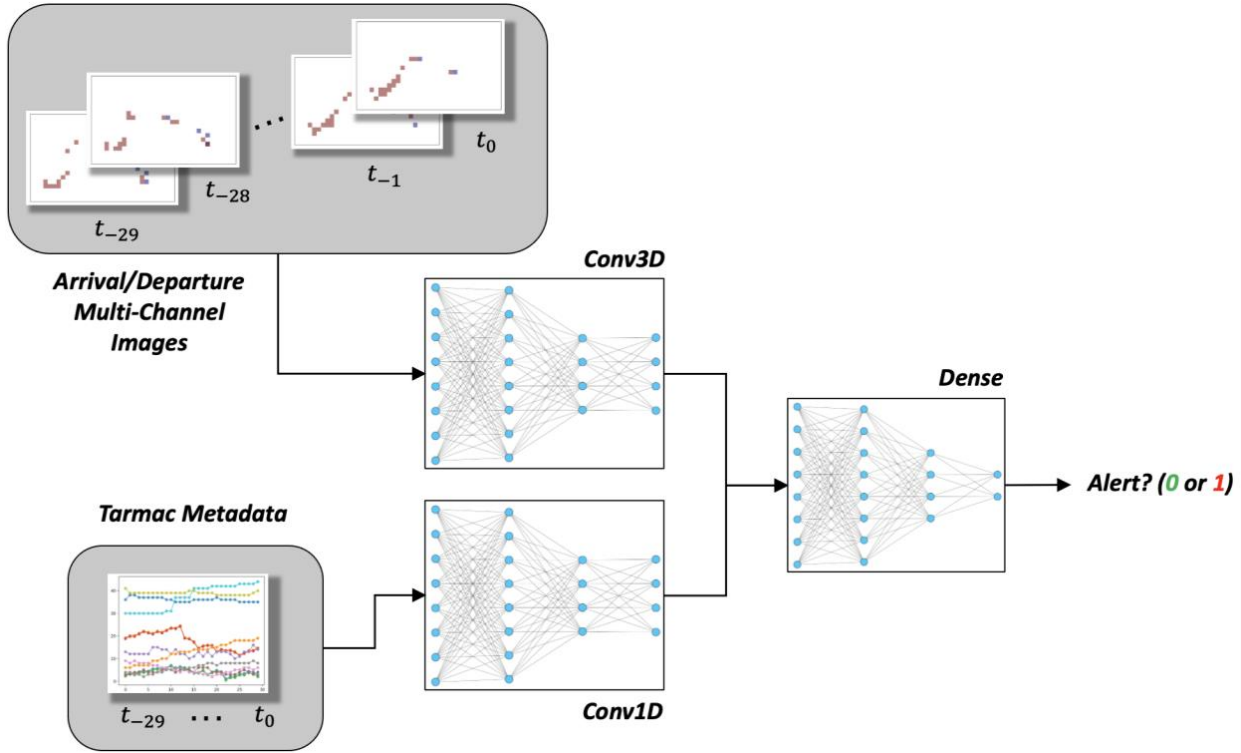

**Figure 3. Illustration of the Neural Network Model for Taxi-Out Time Prediction**

The outputs of the Conv3D and Conv1D layers are flattened and concatenated, and then passed through an additional sequence of dense layers. The final output of the network is a 2D vector with non-negative entries that sum to one and represent the predicted probability of no alert (first component) and an alert (second component). Note that an alert probability greater than 0.5 represents the model's nominal belief that the taxi-out time will exceed the threshold within the next hour.

## V. Model Training and Results

A dataset consisting of: 1) tarmac input feature sequences (the inputs), and 2) average taxi-out time outcomes over the next one hour (the output) was constructed from SWIM downloads collected between 2015 and 2019, inclusive. While a complete discussion of data preparation is beyond the scope of this report, a few details are worth mentioning. First, multiple samples are generated for each historical day via the following process:
   a) Starting at t_0=0Z, define the length-30 tarmac feature inputs over the next 30 minutes, i.e., between t_0=0Z and t_1=0Z+30 minutes;
   b) If the average taxi-out time crossed the threshold in the next one-hour – i.e., between t_1=0Z+30 minutes and t_2=0Z+90 minutes – then mark the sample as a positive; otherwise, make it as a negative;
   c) Increment t_0 by 10 minutes and repeat steps (a) – (c) until the end of the day.

This process yielded between 130,000 and 170,000 samples at each airport of interest over the five years between 2015 and 2019.

At each airport, the sample data is partitioned into three subsets: the first 60% (based on a sorted ordering of samples), the next 20%, and the final 20%. The neural network from Figure 3 is trained on the first 60%, on the condition that the next 20% is used to validate the model and trigger early stopping to the training process when validation loss bottoms out.

There were four airports selected for this study, i.e., Newark Liberty International Airport (EWR), John F. Kennedy International Airport (JFK), LaGuardia Airport (LGA), and Philadelphia International Airport (PHL). A model was trained for each of the airports. Figure 4 contains a breakdown of predictive performance on the held-out test set at these four airports. Note that TN is the number of true-negatives, FP is the number of false-positives, FN is the number of false-negatives, and TP is the number of true-positives. The associated true-negative and true-positive rates are included in parentheses following the true-negative and true-positive counts, respectively. For example, the true-negative and true-positive rates at LGA are both roughly 80%.

The results in Figure 4 could be improved with additional work. Most notably, the current features are not exhaustive and new features, such as weather and downstream restrictions, could be added to enhance the model's prediction power. Performance could be further improved by changing the neural network's architecture (e.g., altering the number of layers and hidden units) or exploring different regularization strategies to achieve a more robust fit.

**EWR**

| $N = 31,117$ | Predicted: No | Predicted: Yes | |
|---|---|---|---|
| Actual: No | TN = 17,776 (68%) | FP = 8,308 | 26,084 |
| Actual: Yes | FN = 1,601 | TP = 3,432 (68%) | 5,033 |
| | 19,377 | 11,740 | |

**JFK**

| $N = 32,864$ | Predicted: No | Predicted: Yes | |
|---|---|---|---|
| Actual: No | TN = 23,157 (75%) | FP = 7,695 | 30,852 |
| Actual: Yes | FN = 502 | TP = 1,510 (75%) | 2,012 |
| | 23,659 | 9,205 | |

**LGA**

| $N = 32,551$ | Predicted: No | Predicted: Yes | |
|---|---|---|---|
| Actual: No | TN = 24,334 (80%) | FP = 6,037 | 30,371 |
| Actual: Yes | FN = 434 | TP = 1,746 (80%) | 2,180 |
| | 24,768 | 7,783 | |

**PHL**

| $N = 31,375$ | Predicted: No | Predicted: Yes | |
|---|---|---|---|
| Actual: No | TN = 21,452 (75%) | FP = 7,206 | 28,658 |
| Actual: Yes | FN = 683 | TP = 2,034 (75%) | 2,717 |
| | 22,135 | 9,240 | |

**Figure 4. Performance on the Held-Out Test Set at EWR, JFK, LGA, and PHL**

Note that during neural network training, samples are weighted to strike a balance between false-positives (i.e., triggering an alert when not warranted) and false-negatives (i.e., failing to trigger an alert when warranted). To enable further tuning of this tradeoff, a subsequent Extreme Gradient Boosting (XGBoost) [8] model is fit against an intermediate output of the neural network. In particular, we extract the output of the first dense layer, which is applied after the Conv1D and Conv3D outputs are concatenated, and train an XGBoost model to maximize its predictive performance across a range of classification thresholds. In our case, a classification threshold $0<\tau<1$ predicts an alert when the probability p returned by the XGBoost model satisfies $p>\tau$, and predicts no alert when $p\leq\tau$. The XGBoost model is trained on the first 80% of data, and the final 20% of data is used to perform a final evaluation of each classification threshold on a continuum of $0<\tau<1$.

In practice, the parameter $\tau$ should be chosen at each airport based on the operational costs of false positives versus false negatives. To select $\tau$ for this paper, we simply choose that the false-positive and false-negative rates are roughly equal. Figure 5 includes receiver operating characteristic (ROC) curves for final taxi-out alerting models trained at the four airports. The ROC curves were constructed by evaluating the false positive rates (x-axis) and true positive rates (y-axis) for each trained model against the last 20% of sample data – unseen during training – over a range of thresholds $\tau$. The dashed black line in each plot represents a baseline model with no skill. The blue curve represents our model's performance over the range $0<\tau<1$. Improvements to the true positive rate coincide with increases to the false positive rate. The right choice of $\tau$ to use in deployment represents an operational tradeoff and should be informed by discussion with traffic managers. For reference, we annotate Figure 5 with a single point that represents the threshold $\tau$ at which false positive and true positive rates are roughly equal.

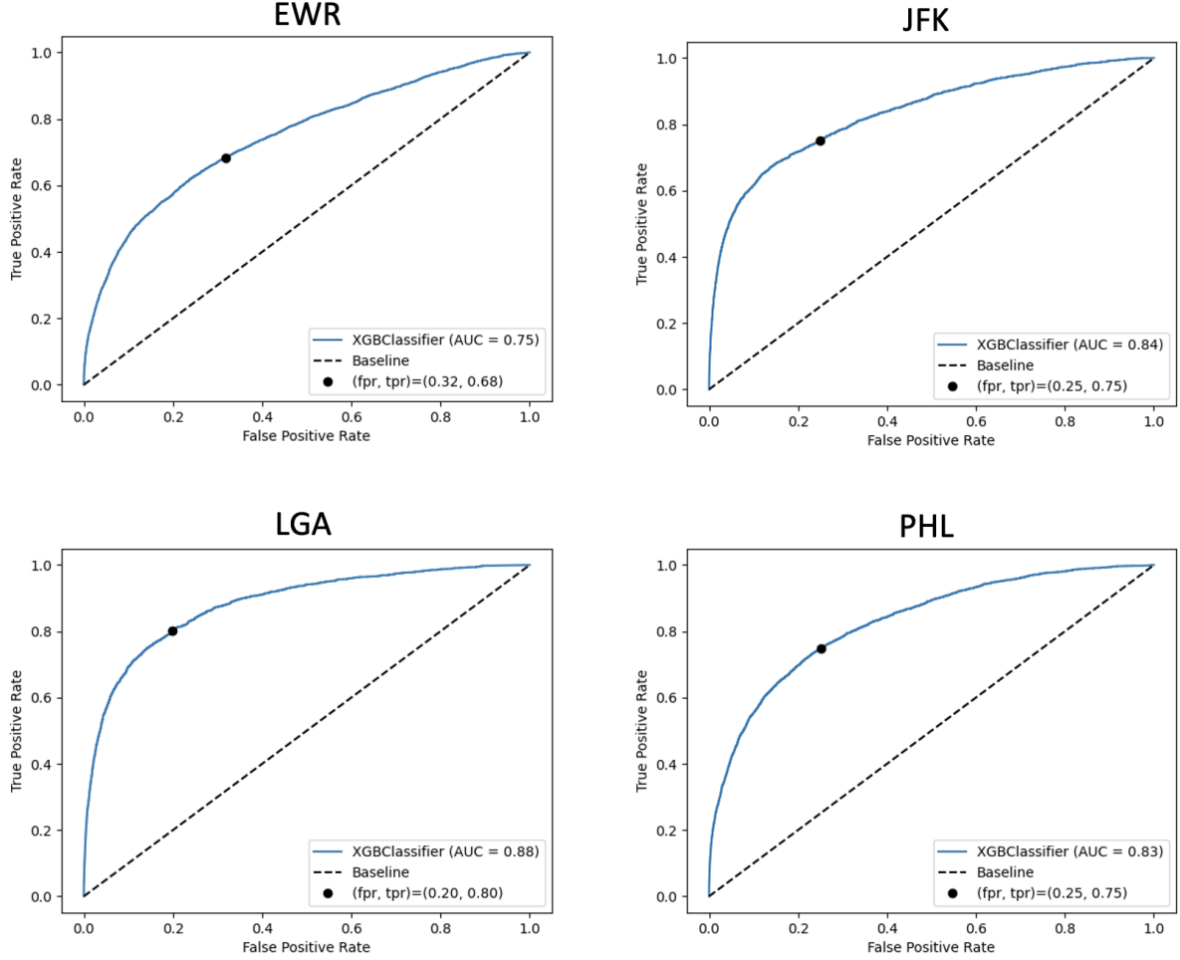

**Figure 5. ROC Curves for Final Taxi-Out Alerting Models Trained at Five Airports in the NAS**

## VI. Exploration of Artificial Intelligence Result Explainability

Explainable AI methods/tools aim to act as a bridge between a black-box ML model output to a more human-interpretable outcome. They help build a trustful relationship between human and ML models when the human experts have the ability to observe and reason the actions being taken by ML with more clarity. In this paper, a computer vision model tries to predict whether the average taxi-out time at an airport will exceed a pre-defined threshold within the next hour of operations or not. The problem was formulated as a computer vision problem by using the time-series images made from position reports of an airport (e.g., LGA). To develop the AI-based prediction, the input images over a 30-minute timeframe are fed to a CNN model where the size of the images are downsized through multiple convolutional layers up to a level that the model can decide if a taxi-out time threshold is met or not.

While humans can anecdotally discern the relationship between runways and the departing flights based on years of operational experience, the CNN model can serve as an assistant function by objectively learning from a diverse and large amount of data to scale up that experience. To help human experts understand what the prediction model learns and gain trust, we use Grad-CAM [9], a popular technique for creating a class-specific heatmap based on a particular input image, to show the region of interest being noticed by the CNN model. Figure 6 shows the departing (red pixels) and arriving (blue pixels) flights at LGA airport for three different timeframes. As can be seen, the Grad-CAM heatmaps show that the ML model reacts to the departing flights (red pixels) and tracks them to make a prediction. For example, Figure 6(a) shows the accumulation of the departing flights near Runway 13 waiting to take off, and the corresponding Grad-CAM heatmap in Figure 6(d) reveals the region of interest to the CNN model. Another example is shown in Figure 6(b), where the departing flights are more evenly distributed over the taxiways, and the corresponding Grad-CAM heatmap shifts its focus from the runway end to a larger area along the taxiway.

Another example is provided here, where the departing flights take off from Runway 31 due to wind condition, as depicted in Figure 6(c). Figure 6(f) shows that the CNN model is now reacting to the departing flights near Runway 31.

The visualization of the trained CNN model at different processing steps (neural layer) helps humans trust the model performance, as it provides a human-interpretable picture of CNN's reaction to the inputs. These examples show that the CNN model is not biased on one specific region or unintended input, and that it can detect different patterns of the departing flights on the airport surface.

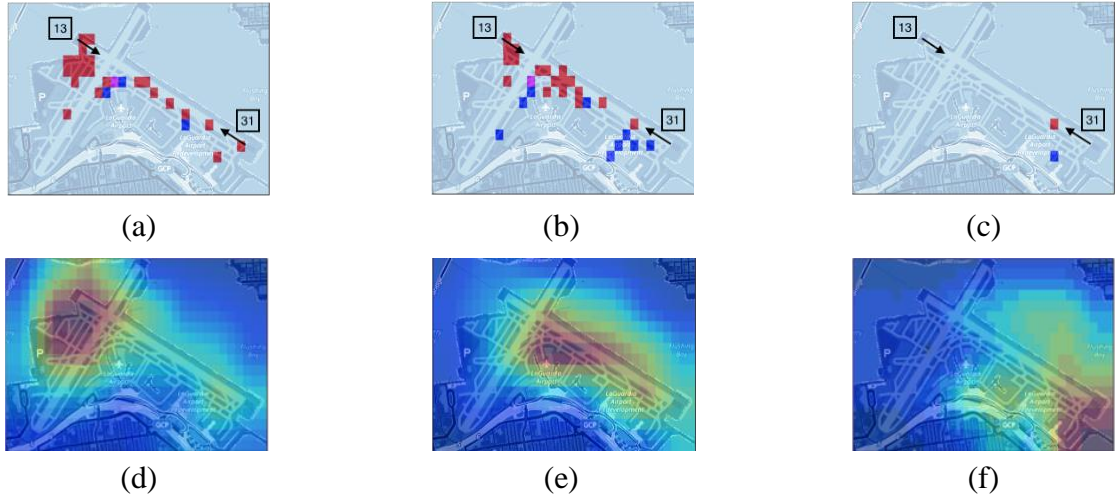

**Figure 6. Arriving and Departing Flights at LGA and Their Corresponding Grad-CAM Heatmaps**

## VII.  Summary

In this paper, we formulated the taxi-out prediction problem, described a novel computer vision modelling approach, and presented initial results. Next steps will include:

- Introducing new features, such as weather and downstream restrictions to improve predictive performance.
- Further tuning the neural network model to optimize predictive performance.
- Eliciting feedback from subject matter experts to improve the operational value of model outputs and visual aids.
- Investigating explainable AI to further enhance interpretability of model outputs.

# References

[1] R. Shumsky, Dynamic Statistical models for the prediction of aircraft take-off times, Cambridge, MA: Operations Research Center, MIT, 1995.

[2] H. Idris, J. Clarke, R. Bhuva and L. Kang, "Queuing Model for Taxi-Out Time Estimation," Air Traffic Control Quarterly, September 11, 2001.

[3] H. Lee, W. Malik and Y. Jung, "Taxi-out time prediction for departures at Charlotte airport using machine learning techniques," in 16th AIAA Aviation Technology, Integration, and Operations Conference, 2016.

[4] F. Herrema, R. Curran, H. Visser, D. Huet and R. Lacote, "Taxi-out time prediction model at Charles de Gaulle Airport," Journal of Aerospace Information Systems, vol. 15, no. 3, pp. 120-130, 2018.

[5] J. Yin, Y. Hu, Y. Ma, Y. Xu, K. Han and D. Chen, "Machine learning techniques for taxi-out time prediction with a macroscopic network topology," in IEEE/AIAA 37th Digital Avionics Systems Conference (DASC), 2018.

[6] J. Legge and B. Levy, "Departure Taxi Time Predictions using ASDE-X Surveillance Data," in 8th AIAA/ATIO, Anchorage, AK.

[7] A. Srivastava, "Improving departure taxi time predictions using ASDE-X surveillance data," in IEEE/AIAA 30th Digital Avionics Systems Conference, 2011.

[8] T. Chen and C. Guestrin, "Xgboost: A scalable tree boosting system," in Proceedings of the 22nd ACM SIGKDD International Conference on Knowledge Discovery and Data Mining, August, 2016.

[9] R. Selvaraju, M. Cogswell, A. Das, R. Vedantam, D. Parikh and D. Batra, "Grad-CAM: Visual Explanations from Deep Networks via Gradient-based Localization," arXiv:1610.02391, 2016.